\documentclass{aptpub}

\usepackage{amsmath,amssymb,amsfonts}
\usepackage{algorithmic}
\usepackage{slashbox}
\usepackage{float}
\usepackage[table, usenames, dvipsnames]{xcolor}
\usepackage[parfill]{parskip}
\usepackage{pifont}%
\usepackage{slashbox}
\usepackage[final]{graphicx}
\usepackage{bm}
\usepackage{textcomp}
\usepackage{dsfont}
\usepackage{subcaption}
\usepackage{grffile}
\usepackage[final]{pdfpages}
\usepackage{pdflscape}
\usepackage{tcolorbox}
\newcounter{magicrownumbers}
\newcommand\rownumber{\stepcounter{magicrownumbers}\arabic{magicrownumbers}. }
\usepackage{adjustbox}
\usepackage{bbm}
\usepackage[normalem]{ulem}
\usepackage{multirow}
\usepackage[misc]{ifsym}
\usepackage{hhline}
\usepackage{array}
\usepackage{makecell}
\usepackage{pifont}
\usepackage{csquotes}
\usepackage{xparse}
\usepackage{xifthen}
\usepackage[pagebackref=true,breaklinks=true]{hyperref}
\usepackage[numbers]{natbib}
\usepackage{etoolbox}
\makeatletter
\pretocmd{\NAT@citexnum}{\@ifnum{\NAT@ctype>\z@}{\let\NAT@hyper@\relax}{}}{}{}
\makeatother
\makeatletter
\newcommand*{\rom}[1]{\expandafter\@slowromancap\romannumeral #1@}
\makeatother
\usepackage{cleveref}
\usepackage{enumitem}
\usepackage{multirow}
\usepackage{ifsym}
\tcbuselibrary{most}
\usetikzlibrary{calc, decorations.pathreplacing, arrows, positioning}

\usepackage{breqn}

\usepackage{hhline}
\newcommand{\Dep}[2]{\text{Dep} \left (#2 \vert #1 \right )}
\newcommand{\Depi}[2]{\text{Dep} \left (#1, #2 \right )}
\newcommand{\UD}[2]{\text{UD} \left (#1, #2 \right )}
\newcommand{\uo}{expected absolute change in distribution }

\newcolumntype{x}[1]{>{\centering\let\newline\\\arraybackslash\hspace{0pt}}p{#1}}

\newcommand{\cmark}{\ding{51}}%
\newcommand{\xmark}{\ding{55}}%

\newcommand{\fully}{fully\ }

\hypersetup{final = true}
\hypersetup{
    colorlinks = true,
    linkcolor={green!30!black},
    citecolor={blue!70!black},
    urlcolor={blue!80!black}
}

\authornames{G.~Berkelmans, J.~Pries {\it et al.}} %
\shorttitle{BP Dependency Function} %

\begin{document}%

\title{\title{The BP Dependency Function: a Generic Measure of Dependence between Random Variables}} %

\authorone[Centrum Wiskunde \& Informatica (CWI)]{Guus Berkelmans \textsuperscript{\Letter}}
\authortwo[Vrije Universiteit (VU)]{Sandjai Bhulai}
\authorthree[Centrum Wiskunde \& Informatica (CWI)]{Rob van der Mei}
\authorfour[Centrum Wiskunde \& Informatica (CWI)]{Joris Pries \textsuperscript{\Letter}}

\addressone{Department of Stochastics, P.O. Box 94079, 1090 GB Amsterdam, Netherlands} %
\addresstwo{Department of Mathematics, De Boelelaan 1111, 1081 HV Amsterdam, Netherlands} 
\addressthree{Department of Stochastics, Science Park 123, 1098 XG Amsterdam, Netherlands} 
\addressfour{Department of Stochastics, P.O. Box 94079, 1090 GB Amsterdam, Netherlands} 
\emailone{gberkelmans@cwi.nl} %
\emailtwo{s.bhulai@vu.nl}
\emailthree{mei@cwi.nl}
\emailfour{joris.pries@cwi.nl}

\footnote{\hspace*{-14pt}\textsuperscript{\Letter}: Equal contribution and corresponding authors}\par

\begin{abstract}
        Measuring and quantifying dependencies between random variables (RV's) can give critical insights into a data-set. Typical questions are: `Do underlying relationships exist?', `Are some variables redundant?', and `Is some target variable $Y$ highly or weakly dependent on variable $X$?' Interestingly, despite the evident need for a general-purpose measure of dependency between RV's, common practice of data analysis is that most data analysts use the {\it Pearson correlation coefficient} (PCC) to quantify dependence between RV's, while it is well-recognized that the PCC is essentially a measure for {\it linear} dependency only. 
        Although many attempts have been made to define more generic dependency measures, there is yet no consensus on a standard, general-purpose dependency function. In fact, several ideal properties of a dependency function have been proposed, but without much argumentation.
        Motivated by this, in this paper we will discuss and revise the list of desired properties and propose a new dependency function that meets all these requirements. This general-purpose dependency function provides data analysts a powerful means to quantify the level of dependence between variables. To this end, we also provide Python code to determine the dependency function for use in practice. 
\end{abstract}



\section{Introduction}
In as early as 1958, Kruskal~\cite{Kruskal} stated that  
\enquote{There are infinitely many possible measures of association, and it sometimes seems that almost as many have been proposed at one time or another.} Many years later, even more dependency measures have been suggested. Yet, and rather surprisingly, there still does not exist consensus on a general dependency function. Often the statement `$Y$ is dependent on $X$' means that $Y$ is not independent of $X$. However, there are different levels of dependency. For example, RV $Y$ can be fully determined by RV $X$ (i.e., $Y(\omega)=f(X(\omega))$ for all $\omega \in \Omega$ and for a measurable function $f$), or only partially.

But how should we quantify how much $Y$ is dependent on $X$? Intuitively, and assuming that the dependency measure is normalized to the interval [0,1], one would say that if $Y$ is fully determined by $X$ then the dependency of $Y$ w.r.t. $X$ is as strong as possible, and so the dependency measure should be 1. On the other side of the spectrum, if $X$ and $Y$ are independent, then the dependency measure should be 0; and vice versa, it is desirable that dependence 0 {\it implies} that $X$ and $Y$ are stochastically independent. Note that the PCC does not meet these requirements. In fact, many examples exists where $Y$ is fully determined by $X$ while PCC$=0$.

Taking a step back, why is it actually useful to examine dependencies in a dataset? Measuring dependencies between the variables can lead to critical insights, which will lead to improved data analysis. First of all, it can reveal important explanatory relationships. How do certain variables interact? If catching a specific disease is highly dependent on the feature value of variable $X$, research should be done to investigate if this information can be exploited to reduce the number of patients with this disease. For example, if hospitalization time is dependent on a healthy lifestyle, measures can be taken to try to improve the overall fitness of a population. Dependencies can therefore function as an actionable steering rod. It is however important to keep in mind that dependency does not always mean causality. Dependency relations can also occur due to mere coincidence or as a byproduct of another process.

Dependencies can also be used for dimensionality reduction. If $Y$ is highly dependent on $X$, not much information is lost when only $X$ is used in the data-set. In this way, redundant variables or variables that provide little additional information, can be removed to reduce the dimensionality of the data-set. With fewer dimensions, models can be trained more efficiently.

In these situations a dependency function can be very useful. However, finding the proper dependency function can be hard, as many attempts have already been made. In fact, most of us have a `gut feeling' for what a dependency function should entail. To make this feeling more mathematically sound, R\'enyi \cite{Renyi} proposed a list of ideal properties for a dependency function. A long list of follow-up papers (see the references in Table~\ref{tab: summary desirable properties} below) use this list as the basis for a wish list, making only minor changes to it, adding or removing some properties. 

In view of the above, the contribution of this paper is threefold:
\begin{itemize}[noitemsep]
    \item We determine a new list of ideal properties for a dependency function;
    \item We present a new dependency function and show that it fulfills all requirements;
    \item We provide Python code to determine the dependency function for the discrete and continuous case.
\end{itemize}

The remainder of this paper is organized as follows.
In Section~\ref{sec: desired properties}, we summarize which ideal properties have been stated in previous literature. By critically assessing these properties, we derive a new list of ideal properties for a dependency function (see Table~\ref{tab: revised desirable properties}), which lays the foundation for a new search for a general-purpose dependency function. 
In Section~\ref{sec: properties previous methods}, the properties are checked for existing methods, and we conclude that 
there does not yet exist a dependency function that has all desired properties. 
Faced by this, in Section~\ref{sec: new dependency} we define a new dependency function and show in Section~\ref{sec: proof properties new dependency measure} that this function meets all the desired properties. Finally, Section~\ref{sec: discussion and conclusion} outlines the general findings and addresses possible future research opportunities.

\section{Desired properties of a Dependency Function} \label{sec: desired properties}
What properties should an ideal dependency function have? In this section, we summarize previously suggested properties. Often, these characteristics are posed without much argumentation. Therefore, we analyze and discuss which properties are actually ideal and which properties are to be believed not relevant, or even wrong.

In Table~\ref{tab: summary desirable properties} below, a summary is given of (twenty-two) 'ideal properties' found in previous literature, grouped into five different categories. These properties are denoted by \hyperref[tab: summary desirable properties]{\rom{1}.1-22}. From these properties we derive a new set of desirable properties denoted by \hyperref[tab: revised desirable properties]{\rom{2}.1-8}, see Table~\ref{tab: revised desirable properties}. Next, we discuss the properties suggested in previous literature and how the new list is derived from them.

{\bf Desired property \rom{2}.1 (Asymmetry):}\\ 
At first glance, it seems obvious that a dependency function should adhere to property \hyperref[tab: 1.13]{\rom{1}.13} and be symmetric. However, this is a common misconception for the dependency function. $Y$ can be fully dependent on $X$, but this does not mean that $X$ is fully dependent on $Y$. Lancaster~\cite{Lancaster} indirectly touched upon this same point by defining \emph{mutual complete dependence}. First it is stated that $Y$ is \emph{completely dependent} on $X$ if $Y=f(X)$. $X$ and $Y$ are called  \emph{mutually completely dependent} if $X$ is completely dependent on $Y$ and vice versa. Thus, this indirectly shows that dependence should not necessarily be symmetric, otherwise the extra definition would be redundant. In \cite{Lancaster} the following great asymmetric example was given.
\begin{ex}
Let $X\sim \mathcal{U}(0,1)$ be uniformly distributed and let $Y = -1$ if $X\leq \frac{1}{2}$ and $Y= 1$ if $X > \frac{1}{2}.$
\label{example: asymmetric 1}
\end{ex}
Then, $Y$ is \fully dependent on $X$, but not vice versa. To drive the point home even more, we give another asymmetric example. \begin{ex}
    $X$ is uniformly randomly drawn out of $\{1,2,3,4\}$ and ${Y := X \mod 2}$.
        \label{example: asymmetric 2}
\end{ex}

$Y$ is fully dependent on $X$, because given $X$ the value of $Y$ is deterministically known. On the other hand, $X$ is not completely known given $Y$. Note that ${Y=1}$ still leaves the possibility for ${X=1}$ or ${X=3}.$ Thus, when assessing the dependency between variable $X$ and variable $Y$, $Y$ is fully dependent on $X$, whereas $X$ is not fully dependent on $Y$. In other words, ${\Depi{X}{Y} \neq \Depi{Y}{X}}$.

In conclusion, {\it an ideal dependency function should not always be symmetric}. To emphasize this point even further, we change the notation of the dependency function. Instead of $\Depi{X}{Y}$, we will denote $\Dep{X}{Y}$ for how much $Y$ is dependent on $X$. Based by this, property \hyperref[tab: 1.13]{\rom{1}.13} is changed into \hyperref[tab: 2.1]{\rom{2}.1}.

{\bf Desired property \rom{2}.2 (Range):}\\
An ideal dependency function should be scaled to the interval $[0,1]$. Otherwise, it can be very hard to draw meaningful conclusions from a dependency score without a known maximum or minimum. What would a score of 4.23 mean without any information about the possible range? Therefore, property \hyperref[tab: 1.1]{\rom{1}.1} is retained. A special note on the range for the well-known \emph{Pearson's correlation coefficient} \cite{Press}, which is $[-1,1]$: The negative or positive sign denotes the direction of the linear correlation. When examining more complex relationships, it is unclear what `direction' entails. We believe that a dependency function should measure by {\it how much} variable $Y$ is dependent on $X$, and not necessarily in which way.

{\bf Desired property \rom{2}.3 (Independence and dependency $\bm{0}$):}\\
If $Y$ is independent of $X$, it should hold that the dependency achieves the lowest possible value, namely zero. Otherwise, it is vague what a dependency score lower than the dependency between two independent variables means. A major issue of the commonly used \emph{Pearson's correlation coefficient}, is that zero correlation does not imply independence. This makes it complicated to derive conclusions from a correlation score. Furthermore, note that if $Y$ is independent of $X$, it should automatically hold that $X$ is also independent of $Y$. In this case, $X$ and $Y$ are independent, because otherwise some dependency relation should exist.

{\bf Desired property \rom{2}.4 (Functional dependence and dependency $\bm{1}$):}\\
If $Y$ is strictly dependent on $X$ (and thus fully determined by $X$), the highest possible value should be attained. It is otherwise unclear what a higher dependency would mean. However, it is too restrictive to demand that the dependency is only 1 if $Y$ is strictly dependent on $X$. R\'enyi~\cite{Renyi} stated \enquote{It seems at the first sight natural to postulate that $\delta(\xi, \eta) = 1$ only if there is a strict dependence of the mentioned type between $\xi$ and $\eta$, but this condition is rather restrictive, and it is better to leave it out}. Take, for example, $Y\sim \mathcal{U}(-1,1)$ and $X:= Y^2$. Knowing $X$ reduces the infinite set of possible values for $Y$ to only two $\left (\pm\sqrt{X}\right )$, whereas it would reduce to one if $Y$ was fully determined by $X$. It would be very restrictive to enforce $\Dep{X}{Y} < 1$, as there is only an infinitesimal difference compared to the strictly dependent case.

{\bf Desired property \rom{2}.5 (Unambiguity):}\\
Kruskal~\cite{Kruskal} stated \enquote{It is important to recognize that the question `Which single measure of association should I use?,' is often unimportant. There may be no reason why two or more measures should not be used; the point I stress is that, whichever ones are used, they should have clear-cut population interpretations.} It is very important that a dependency score leaves no room for ambiguity. The results should stroke with our natural expectation. Therefore, we introduce a new requirement based on a simple example: suppose we have a number of independent RV's and observe one of these at random. The dependency of each random variable on the observed variable should be equal to the probability it is picked. More formally, let ${Y_1,Y_2,\dots,Y_N,S}$ be independent variables with $S$ a selection variable s.t. ${\mathbb{P}(S=i)=p_i}$ and ${\sum_{i=1}^N p_i=1}$. When $X$ is defined as ${X = \sum_{i = 1}^{N} \mathds{1}_{S = i} \cdot Y_i}$, it should hold that ${\Dep{X}{Y_i} = p_i}$ for all ${i\in \{1,\dots, N\}}$. 

{\bf Desired property \rom{2}.6 (Generally applicable):}\\
Our aim is to find a general dependency function, which we denote by $Dep(X|Y)$. This function must be able to handle all kinds of variables: \emph{continuous}, \emph{discrete}, and \emph{categorical} (even nominal). These types of variables occur frequently in a data-set. A general dependency function should be able to measure the dependency of a categorical variable $Y$ on a continuous variable $X$. Stricter than \hyperref[tab: summary desirable properties]{\rom{1}.9-12}, we want a single dependency function that is applicable to any combination of these variables.

There is one exception to this generality. In the case that $Y$ is almost surely constant it is completely independent as well as completely determined by $X$. Arguing what the value of a dependency function should be in this case is a bit similar to arguing the value of $\frac{0}{0}$. Therefore, we argue that in this case it should be either undefined or return some value that represents the fact that $Y$ is almost surely constant (for example $-1$ since this cannot be normally attained).

{\bf Desired property \rom{2}.7 (Invariance under isomorphisms):}\\
Properties \hyperref[tab: summary desirable properties]{\rom{1}.14-20} discuss when the dependency function should be invariant. Most are only meant for variables with an ordering, as `strictly increasing', 'translation' and 'scaling' are otherwise ill-defined. As the dependency function should be able to handle nominal variables, we assume that the dependency is invariant under isomorphisms, see \hyperref[tab: 2.7]{\rom{2}.7}. Note that this is a stronger assumption than \hyperref[tab: summary desirable properties]{\rom{1}.14-20}. Compare Example~\ref{example: asymmetric 2} with the following example.
\begin{ex}
    Let $X'$ be uniformly randomly drawn out of $\{\circ, \triangle, \square, \lozenge \}$ and $Y' = \clubsuit$ if $X' \in \{\circ, \square\}$ and $Y' = \spadesuit$ if $X' \in \{\triangle, \lozenge\}$.
    \label{example: invariant isomorphisms}
\end{ex}
It should hold that $\Dep{X}{Y} = \Dep{X'}{Y'}$ and $\Dep{Y}{X} = \Dep{Y'}{X'}$, as the relationship between the variables is the same (only altered using isomorphisms).

{\bf Desired property \rom{2}.8 (Non-increasing under functions of $\bm{X}$):}\\
Additionally, $\Dep{X}{Y}$ should not increase if a measurable function $f$ is applied to $X$ since any dependence on $f(X)$ corresponds to a dependence on $X$ (but not necessarily the other way around). The information gained from knowing $X$ can only be reduced, never increased by applying a function.

However, though it might be natural to expect the same for functions applied to $Y$, consider once again Example \ref{example: asymmetric 2} (but with $X$ and $Y$ switched around) and the following 2 functions: $f_1(Y):=Y \mod 2$ and $f_2(Y):= \left \lceil\frac{Y}{2}\right \rceil$. Then $f_1(Y)$ is completely predicted by $X$ and should therefore have a dependency of $1$ while $f_2(Y)$ is independent of $X$ and should therefore have a dependency of $0$. So the dependency should be free to increase or decrease for functions applied to $Y$.

{\bf Exclusion of Pearson's correlation coefficient as a special case:}\\
According to properties \hyperref[tab: summary desirable properties]{\rom{1}.21-22}, when $X$ and $Y$ are normally distributed the dependency function should coincide with or be a function of the \emph{Pearson's correlation coefficient}. However, these properties lack a good argumentation for why this would be ideal. It is not obvious why this would be a necessary condition. Even more, there are many known problems and pitfalls with the correlation coefficient \cite{Embrechts,Janse}, so it seems undesirable to force an ideal dependency function to reduce to a function of the correlation coefficient, when the variables are normally distributed. This is why we leave these properties out.

\begin{center}
    \begin{table*}
\caption{A summary of desirable properties for a dependency function stated in previous literature.}
\label{tab: summary desirable properties}
    \begin{adjustbox}{width=\linewidth,keepaspectratio}
        \begin{tabular}{x{4 cm}p{10.5 cm}p{3 cm}} \hline
            \textbf{Property group} & \textbf{Property} & \textbf{Article(s)}\\ \hline
           \multirow{8}{*}[-1.8em]{\textbf{Range}} & \rom{1}.\rownumber $0\leq \Depi{X}{Y} \leq 1$ & \cite{Renyi,Agarwal,Sugiyama,Granger,Joe,Gretton,Szekely,Reshef,Embrechts} \label{tab: 1.1}\\
            & \rom{1}.\rownumber $\Depi{X}{Y} =0 \Leftarrow X$ and $Y$ are independent & \cite{Joe,Granger,Reshef} \label{tab: 1.2}\\
            & \rom{1}.\rownumber $\Depi{X}{Y} =0 \Rightarrow X$ and $Y$ are independent & \cite{Szekely} \label{tab: 1.3}\\
            & \rom{1}.\rownumber $\Depi{X}{Y} =0 \Leftrightarrow X$ and $Y$ are independent & \cite{Renyi,Agarwal,Sugiyama,Gretton,Mori,Embrechts} \label{tab: 1.4}\\
            & \rom{1}.\rownumber $\Depi{X}{Y} = 1 \Leftrightarrow Y=LX$ with probability 1, where $L$ is a similarity transformation & \cite{Mori} \label{tab: 1.5}\\
            & \rom{1}.\rownumber $\Depi{X}{Y} = 1 \Leftarrow X$ and $Y$ are strictly dependent & \cite{Renyi,Agarwal,Granger,Reshef} \label{tab: 1.6}\\
            & \rom{1}.\rownumber $\Depi{X}{Y} = 1 \Leftrightarrow X$ and $Y$ are comonotonic or countermonotonic & \cite{Embrechts} \label{tab: 1.7}\\
             & \rom{1}.\rownumber $\Depi{X}{Y} = 1 \Leftrightarrow X$ and $Y$ are strictly dependent & \cite{Gretton} \label{tab: 1.8}\\ \hline
            & \rom{1}.\rownumber $\Depi{X}{Y}$ is defined for any $X,Y$ where both are not constant & \cite{Renyi,Gretton,Mori} \label{tab: 1.9}\\
            & \rom{1}.\rownumber Well-defined for both continuous and discrete variables & \cite{Granger}\label{tab: 1.10}\\
            & \rom{1}.\rownumber Defined for both categorical and continuous variables; and for ordinal categorical variables for which there may be underlying continuous variables & \cite{Joe} \label{tab: 1.11}\\
            \multirow{-4}{*}{\textbf{General}}& \rom{1}.\rownumber There is a close relationship between the measure for the continuous variables and the measure for the discretization of the variables & \cite{Joe} \label{tab: 1.12}\\ \hline
            \textbf{Symmetric} & \rom{1}.\rownumber $\Depi{X}{Y} = \Depi{Y}{X}$ & \cite{Renyi,Agarwal,Sugiyama,Reshef,Embrechts}\label{tab: 1.13}\\ \hline
            & \rom{1}.\rownumber $\Depi{f(X)}{g(Y)} = \Depi{X}{Y}$ with $f,g$ strictly monotonic functions & \cite{Agarwal}\label{tab: 1.14}\\
            & \rom{1}.\rownumber $\Depi{f(X)}{Y} = \Depi{X}{Y}$ with $f:\mathbb{R}\to \mathbb{R}$ strictly monotonic on the range of $X$ & \cite{Embrechts}\label{tab: 1.15}\\
            & \rom{1}.\rownumber $\Depi{f(X)}{f(Y)} = \Depi{X}{Y}$ with $f$ continuous and strictly increasing & \cite{Sugiyama,Granger} \label{tab: 1.16}\\
            & \rom{1}.\rownumber $\Depi{f(X)}{g(Y)} = \Depi{X}{Y}$ if $f(\cdot), g(\cdot)$ map the real axis in a one-to-one way onto itself & \cite{Renyi,Joe} \label{tab: 1.17}\\
            & \rom{1}.\rownumber  $\Depi{X}{Y}$ is invariant with respect to all similarity transformations & \cite{Mori} \label{tab: 1.18}\\
            & \rom{1}.\rownumber  $\Depi{X}{Y}$ is invariant with respect to translation and scaling & \cite{Sugiyama} \label{tab: 1.19}\\
                \multirow{-7}{*}[4em]{\bf \shortstack{Applying function\\ to argument}}& \rom{1}.\rownumber  $\Depi{X}{Y}$ is scale invariant & \cite{Szekely} \label{tab: 1.20}\\ \hline
            & \rom{1}.\rownumber $\Depi{X}{Y}$ is a function of the Pearson's correlation if the joint distribution of $X$ and $Y$ is normal & \cite{Agarwal,Granger,Szekely} \label{tab: 1.21}\\
            \multirow{-2}{*}{\bf\shortstack{ Behavior normal\\ distribution}} & \rom{1}.\rownumber $\Depi{X}{Y} = \vert \rho(X,Y)\vert $ if the joint distribution of $X$ and $Y$ is normal, where $\rho$ is the Pearson's correlation & \cite{Renyi,Joe} \label{tab: 1.22}\\
                \hline
        \end{tabular}
    \end{adjustbox}
\end{table*}
\end{center}
\setcounter{magicrownumbers}{0}

\begin{center}
    \begin{table*}
\caption{New list of desirable properties for a dependency function}
\label{tab: revised desirable properties}
    \begin{adjustbox}{width=\linewidth,keepaspectratio}
        \begin{tabular}{cp{12.5 cm}} \hline
            \textbf{Property group} & \textbf{Property}\\ \hline
              \textbf{Asymmetric} & \rom{2}.\rownumber There exist RV's $X,Y$ such that $\Dep{X}{Y} \neq \Dep{Y}{X}$. \label{tab: 2.1}\\ \hline
              &\rom{2}.\rownumber $0 \leq \Dep{X}{Y} \leq 1$ for all RV's $X$ and $Y$. \label{tab: 2.2}\\
            & \rom{2}.\rownumber  $\Dep{X}{Y} =0 \Leftrightarrow X$ and $Y$ are independent. \label{tab: 2.3}\\
            & \rom{2}.\rownumber $\Dep{X}{Y} = 1 \Leftarrow Y$ is strictly dependent on $X$. \label{tab: 2.4}\\
            \multirow{-4}{*}{\textbf{Intuitive}} &  \rom{2}.\rownumber
                If $Y_1,Y_2,\dots,Y_N,S$ independent with $\mathbb{P}(S\in[N])=1$, $\mathbb{P}(S=i)=p_i$ and $X=Y_S$ then $\Dep{X}{Y_i}=p_i$ must hold.  \label{tab: 2.5}\\ \hline
            \textbf{General} & \rom{2}.\rownumber Applicable for any combination of continuous, discrete and categorical RV's $X,Y$, where $Y$ is not a.s. constant. \label{tab: 2.6}\\  \hline
            
              & \rom{2}.\rownumber $\Dep{f(X)}{g(Y)}=\Dep{X}{Y}$ for any isomorphisms $f,g$.    \label{tab: 2.7} \\ 
            \multirow{-2}{*}{\textbf{Functions}}  & \rom{2}.\rownumber $\Dep{f(X)}{Y}\leq\Dep{X}{Y}$ for any measurable function $f$. \label{tab: 2.8} \\ \hline
        \end{tabular}
    \end{adjustbox}
\end{table*}
\end{center}
\setcounter{magicrownumbers}{0}

\section{Assessment of the Desired Properties for Existing Dependency Measures} \label{sec: properties previous methods}
In this section, we assess whether existing dependency functions have the properties listed above. In doing so, we limit this section to the most commonly used dependency measures. Table~\ref{tab: previous methods properties} shows which properties each investigated measure adheres to.

Although the desired properties listed in Table~\ref{tab: revised desirable properties} seem not too restrictive, many dependency measures fail to have many of these properties. One of the most commonly used dependency measures, the \emph{Pearson correlation coefficient}, does not even satisfy any one of the desirable properties. Furthermore, almost all measures are not asymmetric. The one measure that comes closes to fulfilling all requirements, is the \emph{uncertainty coefficient} \cite{Press}.
This is a normalized asymmetric variant of the \emph{mutual information} \cite{Press}, where the discrete variant is defined as \begin{align*}
C_{XY}&= \frac{I(X,Y)}{H(Y)} = \frac{\sum_{x,y}p_{X,Y}(x,y) \log \left (\frac{p_{X,Y}(x,y)}{p_X(x) \cdot p_Y(y)} \right )}{- \sum_{y}p_Y(y) \log (p_Y(y))},   
\end{align*}
where $H(Y)$ is the entropy of $Y$ and $I(X,Y)$ is the mutual information of $X$ and $Y$. Note that we use the following notation ${p_{X}(x)=\mathbb{P}(X=x)}$, ${p_{Y}(y)=\mathbb{P}(Y=y)}$, and ${p_{X,Y}(x,y)=\mathbb{P}(X=x,Y=y)}$ throughout the paper. In addition, for a set $H$ we define ${p_{X}(H)=\mathbb{P}(X\in H)}$ (and similarly for $p_Y$ and $p_{X,Y}$).

However, the \emph{uncertainty coefficient} does not satisfy properties \hyperref[tab: 2.5]{\rom{2}.5} and \hyperref[tab: 2.6]{\rom{2}.6}. For example, if $Y\sim \mathcal{U}(0,1)$ is uniformly drawn, the entropy of $Y$ becomes: \begin{align*}
    H(Y) &= - \int_{0}^{1} f_Y(y)\ln\left(f_Y(y)\right )dy\\
    &= - \int_{0}^{1} 1 \cdot \ln\left(1\right )dy\\
    &= 0.
\end{align*}
Thus, for any $X$, the uncertainty coefficient is now undefined (division by zero). Therefore, the uncertainty coefficient is not as generally applicable as property \hyperref[tab: 2.6]{\rom{2}.5} requires. 

Two other measures that satisfy many (but not all) properties are \emph{mutual dependence} \cite{Agarwal} and \emph{maximal correlation} \cite{Gebelein}. Mutual dependence is defined as the Hellinger distance \cite{Hellinger} $d_h$ between the joint distribution and the product of the marginal distributions, defined as follows (cf. \cite{Agarwal}): 
\begin{align}
    d(X,Y) \triangleq d_h(f_{XY}(x,y), f_X(x)\cdot f_Y(y)). \label{eq: mutual dependence}
\end{align}
Maximal correlation is defined as (cf. \cite{Renyi}): \begin{align}
    S(X, Y)= \sup_{f,g} R(f(X), g(Y)), \label{eq: maximal correlation}
\end{align}
where $R$ is the Pearson correlation coefficient, and where $f,g$ are Borel-measurable functions, such that $R(f(X), g(Y))$ has a sense \cite{Renyi}.

Clearly, Equations~(\ref{eq: mutual dependence}) and (\ref{eq: maximal correlation}) are symmetric. The joint distribution and the product of the marginal distributions does not change by switching $X$ and $Y$. Furthermore, the Pearson correlation coefficient is symmetric, making the maximal correlation also symmetric. Therefore, both measures do not have property \hyperref[tab: 2.1]{\rom{2}.1}  

\begin{center}
    \begin{table*}
\caption{Properties of previous dependencies functions (\xmark = property not satisfied, \cmark = property satisfied)}
\label{tab: previous methods properties}
    \begin{adjustbox}{width=\linewidth,keepaspectratio}
        \begin{tabular}{lcccccccc} \hline
            & \textbf{Asymmetric} & \multicolumn{4}{c}{\textbf{Intuitive}}  & \textbf{General} &   \multicolumn{2}{c}{\textbf{Functions}}  \\ 
             \multirow{-2}{*}{\textbf{Measure}} & \hyperref[tab: 2.1]{\rom{2}.1} & \hyperref[tab: 2.2]{\rom{2}.2} & \hyperref[tab: 2.3]{\rom{2}.3} & \hyperref[tab: 2.4]{\rom{2}.4} & \hyperref[tab: 2.5]{\rom{2}.5} & \hyperref[tab: 2.6]{\rom{2}.6} &
            \hyperref[tab: 2.6]{\rom{2}.7} & \hyperref[tab: 2.7]{\rom{2}.8} \\ \hline
            Pearson correlation coefficient \cite{Press} & \xmark & \xmark & \xmark & \xmark & \xmark & \xmark & \xmark & \xmark \\
            Spearman's rank correlation coefficient \cite{Press}  & \xmark & \xmark & \xmark & \xmark & \xmark & \xmark & \xmark & \xmark \\
            Kendall rank correlation coefficient \cite{Press} & \xmark & \xmark & \xmark & \xmark & \xmark & \xmark & \xmark & \xmark \\
            Mutual information \cite{Press}  & \xmark & \xmark & \cmark & \xmark & \xmark & \cmark & \cmark & \cmark \\
            Uncertainty coefficient \cite{Press} & \cmark & \cmark & \cmark & \cmark & \xmark & \xmark & \cmark & \cmark \\
            Total correlation \cite{Watanabe}  & \xmark & \xmark & \cmark & \xmark & \xmark & \cmark & \cmark & \cmark \\
            Mutual dependence \cite{Agarwal}  & \xmark & \cmark & \cmark & \cmark & \xmark & \cmark & \cmark & \cmark \\
            $\Delta_{L_1}$ \cite{CAPITANI}  & \xmark & \xmark & \cmark & \xmark & \xmark & \cmark & \cmark & \cmark \\
            $\Delta_{SD}$ \cite{CAPITANI}  & \xmark & \xmark & \cmark & \xmark & \xmark & \cmark & \xmark & \xmark \\
            $\Delta_{ST}$ \cite{CAPITANI}  & \xmark & \xmark & \xmark & \xmark & \xmark & \cmark & \xmark & \xmark \\
            Monotone correlation \cite{Kimeldorf}  & \xmark & \cmark & \cmark & \xmark & \xmark & \xmark & \xmark & \xmark \\
            Maximal correlation \cite{Gebelein}  & \xmark & \cmark & \cmark & \cmark & \xmark & \cmark & \cmark & \cmark \\
            Distance correlation \cite{Szekely} & \xmark & \cmark & \cmark & \xmark & \xmark & \xmark & \xmark & \xmark \\ \hline
        \end{tabular}
    \end{adjustbox}
\end{table*}
\end{center}

\section{The Berkelmans-Pries Dependency Function} \label{sec: new dependency}
After devising a new list of ideal properties (see Table~\ref{tab: revised desirable properties}) and showing that these properties are not fulfilled by existing dependency functions (see Table~\ref{tab: previous methods properties}), we will now introduce a new dependency function that will meet all requirements. Throughout, we will refer to this function as the {\it Berkelmans-Pries (BP)} dependency function.

The key question surely is: What is dependency? Although this question deserves an elaborate philosophical study, we believe that measuring the dependency of $Y$ on $X$, is essentially measuring how much the distribution of $Y$ changes on average based on the knowledge of $X$, divided by the maximum possible change. This is the key insight, where the {\it BP} dependency function is based on. To measure this, we first have to determine the difference between the distribution of $Y$ \emph{with} and \emph{without} conditioning on the value of $X$ times the probability that $X$ takes on this value in Section~\ref{sec: \uo}. Secondly, we have to measure what the maximum possible change in probability mass is, which is used to properly scale the dependency function and make it asymmetric (see Section~\ref{sec: max \uo}).

\subsection{Definition \uo} \label{sec: \uo}
We start by measuring the \emph{\uo\!\!} (UD), which is the difference between the distribution of $Y$ \emph{with} and \emph{without} conditioning on the value of $X$ times the probability that $X$ takes on this value. To this end, for two discrete RV's $X$ and $Y$, UD is defined as follows:

\begin{dmath*}
    \UD{X}{Y}\hiderel{=}\sum_{x} p_X(x) \cdot \sum_{y} \left \vert p_{Y\vert X=x}(y)   - p_Y(y) \right\vert .
\end{dmath*}

More explicit formulations of UD for specific combinations of RV's are given in Appendix~\ref{appendix: formulations of UD}. For example, when $X$ and $Y$ remain discrete and take values in $E_X$ and $E_Y$ respectively, it can equivalently be defined as:
\begin{dmath*}
    \UD{X}{Y}\hiderel{=}2\sup_{A\subset E_X \times E_Y} \left \{\sum_{(x,y)\in A}(p_{X,Y}(x,y)-p_X(x) \cdot p_Y(y)) \right \}. 
\end{dmath*}

Similarly, if $X$ and $Y$ are two continuous RV's, UD can be written as
\begin{dmath*}
    \UD{X}{Y}\hiderel{=}\int_{\mathbb{R}}\int_{\mathbb{R}}\vert f_{X,Y}(x,y)-f_X(x)f_Y(y)\vert dy dx,
\end{dmath*}
which is the same as $\Delta_{L_1}$ \cite{CAPITANI}.

In general, for ${X: (\Omega,\mathcal{F},\mu)\rightarrow (E_X,\mathcal{E}(X))}$ and ${Y: (\Omega,\mathcal{F},\mu)\rightarrow (E_Y,\mathcal{E}(Y))}$ UD is defined as
\begin{dmath*}
    \UD{X}{Y} 
		=2\sup_{A\in\mathcal{E}(X)\bigotimes\mathcal{E}(Y)} \left \{ \mu_{(X,Y)}(A)-(\mu_X \hiderel{\times}\mu_Y)(A) \right \},
\end{dmath*}
where ${\mathcal{E}(X)\bigotimes\mathcal{E}(Y)}$ is the $\sigma$-algebra generated by the sets ${C\times D}$ with ${C\in\mathcal{E}(X)}$ and ${D\in\mathcal{E}(Y)}$. Furthermore, $\mu_{(X,Y)}$ denotes the joint probability measure on ${\mathcal{E}(X)\bigotimes\mathcal{E}(Y)}$ and ${\mu_X \times \mu_Y}$ is the product measure.

\subsection{\texorpdfstring{Maximum UD given $Y$}{}
} \label{sec: max \uo}
Next, we have to determine the maximum of UD for a fixed $Y$ in order to scale the dependency function to $[0,1]$. To this end, we prove that for a given $Y$: \begin{dmath*}
X \text{ \fully determines } Y \Rightarrow    \UD{X}{Y} \hiderel{=} \max_{X'} \left \{ \UD{X'}{Y} \right \}. \label{eq: statement maximum \uo}
\end{dmath*}

The full proof for the general case is given in Appendix~\ref{appendix: proof maximum UD achieved}, which uses the upper bound determined in Appendix~\ref{appendix: proof maximality \uo}. However, we will show the discrete case here to give some intuition about the proof.
Let ${C_y=\left \{x\vert p_{X,Y}(x,y)\geq p_X(x)\cdot p_Y(y)\right \}}$, then
\begin{dmath*}
\UD{X}{Y}=2\sum_y \left (p_{X,Y}(C_y\times\{y\})-p_X(C_y) \cdot p_Y(y) \right )\\
\leq 2\sum_y \left (\min\left\{p_X(C_y), p_Y(Y)\right\}-p_X(C_y)\cdot p_Y(y) \right ) \\
= 2\sum_y \left (\min \left \{p_X(C_y) \hiderel{\cdot} (1-p_Y(y)), (1-p_X(C_y)) \hiderel{\cdot} p_Y(y)\right \}\right )\\
    \leq 2\sum_y \left (p_Y(y) \cdot (1-p_Y(y))\right )\\
    =2\sum_y \left (p_Y(y)-p_Y(y)^2\right )\\
    =2\left (1-\sum_y p_Y(y)^2 \right ),\\
\end{dmath*}
with equality iff both inequalities are equalities. Which occurs iff $p_{X,Y} \left (C_y\times\{y\} \right )=p_X(C_y)=p_Y(y)$ for all $y$. So we have equality when for all $y$ the set $C_y$ has the property that $x\in C_y$ iff $Y=y$. Or equivalently $Y=f(X)$ for some function $f$. Thus,
\begin{dmath*}
    UD(X,Y)\leq 2\left(1-\sum_y p_Y(y)^2 \right ),
\end{dmath*}
with equality iff $Y=f(X)$ for some function $f$.
     
Note that this holds for every $X$ that \fully determines $Y$. In particular, for ${X:=Y}$ it now follows that
\begin{align*}
    \UD{Y}{Y} = \max_{X'}\left \{\UD{X'}{Y}\right \}=2\cdot (1-\sum_y p_Y(y)^2).
\end{align*}

\subsection{Definition Berkelmans-Pries Dependency Function} \label{sec: definition new dependency}
Finally, we can define the {\it BP} dependency function to measure how much $Y$ is dependent on $X$, by
\begin{align*}
\def\arraystretch{1.5}
    \Dep{X}{Y}= \left \{ \begin{array}{cl}
       \frac{\UD{X}{Y}}{\UD{Y}{Y}}  & \text{if $Y$ is not a.s. constant,}  \\
        \text{undefined}    & \text{if $Y$ is a.s. constant.} \\
    \end{array}\right . \label{eq: definition new dependency function with exception}
\end{align*}
This is the difference between the distribution of $Y$ \emph{with} and \emph{without} conditioning on the value of $X$ times the probability that $X$ takes on this value divided by the largest possible difference for an arbitrary $X'$. Note that $\UD{Y}{Y}=0$ if and only if $Y$ is almost surely constant (see Appendix~\ref{appendix: proof maximum UD achieved}), which leads to division by zero. However, we previously argued in Section~\ref{sec: desired properties} that if $Y$ is almost surely constant, it is completely independent as well as completely determined by $X$. It should therefore be undefined.

\section{Properties of the Berkelmans-Pries Dependency Function} \label{sec: proof properties new dependency measure}
Next, we show that our new {\it BP} dependency function satisfies all requirements from Table~\ref{tab: revised desirable properties}. To this end, we use properties of UD (see Appendix~\ref{appendix: properties \uo}) to derive properties \hyperref[tab: revised desirable properties]{\rom{2}.1-8}.

{\bf Property \rom{2}.1 (Asymmetry):}
It holds for Example~\ref{example: asymmetric 1} that ${\UD{X}{Y} = 1}$, \\ ${\UD{X}{X} = 2}$, and ${\UD{Y}{Y} = 1}.$ Thus, \begin{align*}
    \Dep{X}{Y} &= \frac{\UD{X}{Y}}{\UD{Y}{Y}} = 1, \\
    \Dep{Y}{X} &= \frac{\UD{X}{Y}}{\UD{X}{X}} = \frac{1}{2}.
\end{align*}
Therefore, we see that ${\Dep{X}{Y} \neq \Dep{Y}{X}}$ for this example, thus making the {\it BP} dependency asymmetric.

{\bf Property \rom{2}.2 (Range):} In Appendix~\ref{appendix: zero iff independent \uo}, we show that for every $X,Y$ it holds that ${\UD{X}{Y} \geq 0}$. Furthermore, in Appendix~\ref{appendix: proof maximality \uo} we prove that $\sup_{X'}\left \{\UD{X}{Y}\right \}\leq 2\left (1-\sum_{y\in d_Y} \mu_Y(\{y\})^2\right ).$ In Appendix~\ref{appendix: proof maximum UD achieved} we show for almost all cases that this bound is tight for $\UD{Y}{Y}$. Thus, it must hold that ${0\leq \UD{X}{Y}\leq \UD{Y}{Y}}$ and it then immediately follows that ${0\leq \Dep{X}{Y}\leq 1}$. The result for the highly specific remaining cases remains an open problem, the formulation (and conjecture) of which can be found in the Appendix~\ref{appendix: proof maximum UD achieved}.

{\bf Property \rom{2}.3 (Independence and dependency $\bm{0}$):}
In Appendix~\ref{appendix: zero iff independent \uo}, we prove that \begin{align*}
    \UD{X}{Y} = 0 \Leftrightarrow X \text{ and } Y \text{ are independent}.
\end{align*}
Furthermore, note that ${\Dep{X}{Y}=0}$ if and only if ${\UD{X}{Y} = 0}$. Thus, \begin{align*}
    \Dep{X}{Y} = 0 \Leftrightarrow X \text{ and } Y \text{ are independent}.
\end{align*}

{\bf Property \rom{2}.4 (Functional dependence and dependency $\bm{1}$):} In Section~\ref{appendix: proof maximum UD achieved}, we show that if $X$ \fully determines $Y$ it holds that ${\UD{X}{Y} = \max_{X'}\left \{ \UD{X'}{Y}\right \} }$, in almost all cases. This holds in particular for $X:=Y$. Thus, if $X$ \fully determines $Y$ it follows that
\begin{align*}
    \Dep{X}{Y} = \frac{\UD{X}{Y}}{\UD{Y}{Y}} = \frac{\max_{X'}\UD{X'}{Y}}{\max_{X'}\UD{X'}{Y}} = 1.
\end{align*}
The result for the highly specific remaining cases remains an open problem, the formulation (and conjecture) of which can be found in the Section~\ref{appendix: proof maximum UD achieved}.

{\bf Property \rom{2}.5 (Unambiguity):}
Let $\mathcal{E}$ denote the $\sigma$-algebra, where ${Y_1,Y_2,\dots,Y_N}$ are defined. By definition, it holds that ${\mathbb{P}(X=x)=\sum_j\mathbb{P}(Y_j=x) \cdot \mathbb{P}(S=j)}$, so for all ${i\in \{1, \dots N\}}$
\begin{dmath*}
\UD{X}{Y_i}=2\sup_{A\in\mathcal{E}\times\mathcal{E}}\left \{\sum_{(x,y)\in A}\left (\mathbb{P}(X\hiderel{=}x,Y_i\hiderel{=}y)-\mathbb{P}(X\hiderel{=}x)\mathbb{P}(Y_i\hiderel{=}y)\right ) \right \}
=2\sup_{A\in\mathcal{E}\times\mathcal{E}} \left \{  \sum_{(x,y)\in A}\left ( \sum_j\mathbb{P}(Y_S\hiderel{=}x,Y_i\hiderel{=}y,S\hiderel{=}j)-\mathbb{P}(X\hiderel{=}x)\mathbb{P}(Y_i\hiderel{=}y)\right ) \right \}
=2\sup_{A\in\mathcal{E}\times\mathcal{E}} \left \{  \sum_{(x,y)\in A}\left (\sum_{j\neq i}\mathbb{P}(Y_S\hiderel{=}x,Y_i\hiderel{=}y,S\hiderel{=}j)+\mathbb{P}(Y_S\hiderel{=}x,Y_i\hiderel{=}y,S\hiderel{=}i)-\mathbb{P}(X\hiderel{=}x)\mathbb{P}(Y_i\hiderel{=}y) \right )\right\}
=2\sup_{A\in\mathcal{E}\times\mathcal{E}} \left \{  \sum_{(x,y)\in A}\left ( \sum_{j\neq i}\mathbb{P}(Y_j\hiderel{=}x)\mathbb{P}(Y_i\hiderel{=}y)\mathbb{P}(S\hiderel{=}j)+\mathbb{P}(Y_i\hiderel{=}x,Y_i\hiderel{=}y)\mathbb{P}(S\hiderel{=}i)-\sum_j\mathbb{P}(Y_j\hiderel{=}x)\mathbb{P}(S\hiderel{=}j)\mathbb{P}(Y_i\hiderel{=}y) \right ) \right\}
=2\sup_{A\in\mathcal{E}\times\mathcal{E}} \left\{ \sum_{(x,y)\in A}\left ( p_i\mathbb{P}(Y_i\hiderel{=}x,Y_i\hiderel{=}y)-p_i\mathbb{P}(Y_i\hiderel{=}x)\mathbb{P}(Y_i\hiderel{=}y)\right ) \right\}
=p_i\UD{Y_i}{Y_i}.
\end{dmath*}
This leads to
\begin{align*}
    \Dep{X}{Y_i}&=\frac{\UD{X}{Y_i}}{\UD{Y_i}{Y_i}} = \frac{p_i\UD{Y_i}{Y_i}}{\UD{Y_i}{Y_i}}=p_i.
\end{align*}
Therefore, we can conclude that property \hyperref[tab: 2.5]{\rom{2}.5} holds.

{\bf Property \rom{2}.6 (Generally applicable):} The {\it BP} dependency measure can be applied for any combination of continuous, discrete and categorical variables. It can handle arbitrary many RV's as input by combining them. Thus, the {\it BP} dependency function is generally applicable.

{\bf Property \rom{2}.7 (Invariance under isomorphisms):}
In Appendix~\ref{appendix: invariant isomorphisms \uo}, we show that UD is invariant under isomorphisms. In other words, for any isomorphisms $f,g$ it holds that
\begin{align*}
    \UD{X}{Y} = \UD{f(X)}{g(Y)}.
\end{align*}
It follows for the {\it BP} dependency measure that

\begin{dmath*}
\Dep{f(X)}{g(Y)}=\frac{\UD{f(X)}{g(Y)}}{\UD{g(Y)}{g(Y)}}=\frac{\UD{X}{Y}}{\UD{Y}{Y}}\\
=\Dep{X}{Y},
\end{dmath*}
thus Property \hyperref[tab: 2.7]{\rom{2}.7} is satisfied.

{\bf Desired property \rom{2}.8 (Non-increasing under functions of $\bm{X}$):}
In Appendix~\ref{appendix: never increase \uo}, we prove that transforming $X$ or $Y$ using a measurable function does not increase UD. In other words, for any measurable function $f$, it holds that \begin{align*}
    \UD{f(X)}{Y} \leq \UD{X}{Y}.
\end{align*}
Consequently, Property \hyperref[tab: 2.8]{\rom{2}.8} holds for the {\it BP} dependency function, as \begin{dmath*}
\Dep{f(X)}{Y}=\frac{\UD{f(X)}{Y}}{\UD{Y}{Y}}\leq\frac{\UD{X}{Y}}{\UD{Y}{Y}}=\Dep{X}{Y}.
\end{dmath*}

\section{Discussion and Further Research} \label{sec: discussion and conclusion}

Motivated by the need to measure and quantify the level dependence between random variables, we have proposed a general-purpose dependency function. The function  meets an extensive list of important and desired properties, and can be viewed as a powerful alternative to the classical PCC, which is often used by data analysts today.

Whilst it is recommended to use our new dependency function, it is important to understand the limitations and potential pitfalls of the new dependency function. Below we elaborate on these aspects.

The underlying probability density function of a RV is often unknown in practice; instead, a set of outcomes is observed. These samples can then be used (in a simple manner) to approximate any discrete distribution. However, this is generally not the case for continuous variables. There are mainly two categories for dealing with continuous variables: either (1) the observed samples are combined using kernel functions into a continuous function (\emph{kernel density estimation} \cite{Gramacki}), or (2) the continuous variable is reduced to a discrete variable using \emph{data binning}. The new dependency measure can be applied thereafter.

\begin{figure}[H]
    \centering
    \begin{subfigure}{.47 \textwidth}
        \centering
            \includegraphics[keepaspectratio, width = \linewidth]{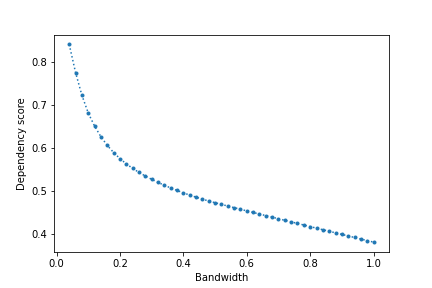}
        \caption{\small \textbf{Kernel density estimation}}
         \label{fig: bandwidth plot}
    \end{subfigure}%
    \hfill
    \begin{subfigure}{.47 \textwidth}
        \centering
            \includegraphics[keepaspectratio, width = \linewidth]{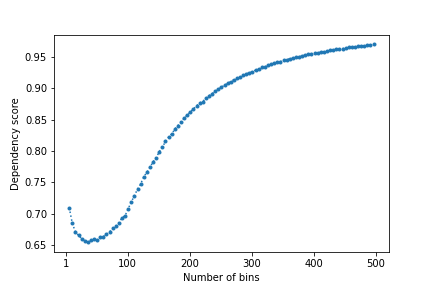}
            \caption{\small \textbf{Data binning}}
         \label{fig: bin plot}
    \end{subfigure}
    
    \caption{Influence of chosen bandwidth (a) / number of bins (b) on the dependency score $\Dep{X}{Y}$ with $5{,}000$ samples of ${X\sim \mathcal{U}(0,1)}$ and ${Y = X + \epsilon}$ with ${\epsilon \sim \mathcal{N}(0,0.1).}$ }
\end{figure}

A main issue is that the dependency measure is dependent of parameter choices of either \emph{kernel density estimation} or \emph{data binning}. To illustrate this, we conduct the following experiment: Let ${X\sim\mathcal{U}(0,1)}$ and define ${Y = X + \epsilon}$ with ${\epsilon\sim\mathcal{N}(0,0.1)}$. Next, we draw ${5{,}000}$ samples of $X$ and $\epsilon$ and determine each corresponding $Y$. For \emph{kernel density estimation}, we use Gaussian kernels with constant bandwidth. The result of varying the bandwidth on the dependency score can be seen in Figure~\ref{fig: bandwidth plot}. With \emph{data binning}, both $X$ and $Y$ are binned using bins with fixed size. Increasing or decreasing the number of bins changes the size of the bins. The impact of changing the number of bins on the dependency score, can be seen in Figure~\ref{fig: bin plot}.

The main observation from Figures~\ref{fig: bandwidth plot} and~\ref{fig: bin plot} is that the selection of the parameters is important. In the case of the \emph{kernel density estimation}, we see the traditional {\it trade-off} between {\it over-fitting} when the bandwidth is too small and {\it under-fitting} when the bandwidth is too large. On the other hand, with \emph{data binning}, we see different behaviour: Having too few bins seems to overestimate the dependency score and as bins increase the estimator of the dependency score decreases up to a certain point, where-after it starts increasing again. The bottom of the curve seems to be marginally higher than the true dependency score of 0.621. 

This observation raises a range of interesting questions for future research. For example, are the dependency scores estimated by binning consistently higher than the true dependency? Is there a correction that can be applied to get an unbiased estimator? Is the minimum of this curve an asymptotically consistent estimator? Which binning algorithms give the closest approximation of the true dependency? 

An interesting observation, with respect to kernel density estimation, is that it appears that at a bandwidth of 0.1 the estimator of the dependency score is close to the true dependency score of approximately 0.621. However, this parameter choice could only be made if the underlying probability process was known {\it a priori}. 

Yet, there is another challenge with kernel density estimation, when $X$ consists of many variables or feature values. Each time $Y$ is conditioned on a different value of $X$, either the density needs to be estimated again or the estimation of the joint distribution needs to be integrated. Both can rapidly become very time-consuming. When using data binning, it suffices to bin the data once. Furthermore, no integration is required making it much faster. Therefore, our current recommendation would be to bin the data and not use kernel density estimation.

\appendix
\section{}
\subsection*{Notation}
The following general notation is used throughout this appendix. Let $X:(\Omega,\mathcal{F},\mathbb{P})\to (E_X,\mathcal{E}_X)$ and ${Y:(\Omega,\mathcal{F},\mathbb{P})\to (E_Y,\mathcal{E}_Y)}$ be RV's. Secondly, let ${\mu_{X}(A)=\mathbb{P}(X^{-1}(A))}$, ${\mu_{Y}(A)=\mathbb{P}(Y^{-1}(A))}$ be measures induced by $X$ and $Y$ on $(E_X,\mathcal{E}_X)$ and $(E_Y,\mathcal{E}_Y)$ respectively. Furthermore, ${\mu_{X,Y}(A)=\mathbb{P}(\{\omega\in\Omega\vert (X(\omega),Y(\omega)\})\in A)}$ is the joint measure and ${\mu_X \times \mu_Y}$ is the product measure on ${(E_X\times E_Y,\mathcal{E}_X\bigotimes\mathcal{E}_Y)}$ generated by \\ $(\mu_X\times\mu_Y)(A\times B)=\mu_X(A)\mu_Y(B)$.\\
\subsection{Formulations of UD} \label{appendix: formulations of UD}
In this appendix, we give multiple formulations of the \emph{\uo \!\!} (UD). Depending on the type of RV's, these formulations can be used.\\

\subsubsection{\bf General case}
For any $X$, $Y$ the UD is defined as
\begin{dmath}
    \UD{X}{Y} \hiderel{=} \sup_{A\in\mathcal{E}(X)\bigotimes\mathcal{E}(Y)} \left \{ \mu_{(X,Y)}(A)-(\mu_X \hiderel{\times}\mu_Y)(A) \right \}
		+\sup_{B\in\mathcal{E}(X)\bigotimes\mathcal{E}(Y)} \left \{ (\mu_X \hiderel{\times}\mu_Y)(B)-\mu_{(X,Y)}(B) \right \} \\
		=2\sup_{A\in\mathcal{E}(X)\bigotimes\mathcal{E}(Y)} \left \{ \mu_{(X,Y)}(A)-(\mu_X \hiderel{\times}\mu_Y)(A) \right \}, \label{eq: general definition \uo}
\end{dmath}

\subsubsection{\bf Discrete RV's only}
When $X,Y$ are discrete RV's, Equation~\ref{eq: general definition \uo} simplifies into
\begin{dmath*}
    \UD{X}{Y}\hiderel{=}\sum_{x,y} \left \vert p_{X,Y}(x,y)-p_X(x)\cdot p_Y(y) \right \vert, \label{eq: discrete notation 2}
\end{dmath*}
or equivalently,
\begin{dmath*}
    \UD{X}{Y}\hiderel{=}\sum_{x} p_X(x) \cdot \sum_{y} \left \vert p_{Y\vert X=x}(y)   - p_Y(y) \right\vert \label{eq: discrete \uo}.
\end{dmath*}

Similarly, when $X$ and $Y$ take values in $E_X$ and $E_Y$ respectively, Equation~\ref{eq: general definition \uo} becomes
\begin{dmath*}
    \UD{X}{Y}\\
    \hiderel{=}\sup_{A\subset E_X \times E_Y} \left \{\sum_{(x,y)\in A}(p_{X,Y}(x,y)-p_X(x)p_Y(y)) \right \}\\
    \hiderel{+}\sup_{A\subset E_X \times E_Y} \left \{\sum_{(x,y)\in A}(p_X(x)p_Y(y)-p_{X,Y}(x,y)) \right\}\\
    \hiderel{=}2\sup_{A\subset E_X \times E_Y} \left \{\sum_{(x,y)\in A}(p_{X,Y}(x,y)-p_X(x)p_Y(y)) \right \}. \label{eq: discrete notation 3}
\end{dmath*}

\subsubsection{\bf Continuous RV's only}
When $X,Y$ are continuous RV's, Equation~\ref{eq: general definition \uo} becomes:
\begin{dmath*}
    \UD{X}{Y}\hiderel{=}\int_{\mathbb{R}}\int_{\mathbb{R}}\vert f_{X,Y}(x,y)-f_X(x)f_Y(y)\vert dy dx,
\end{dmath*}
or equivalently
\begin{dmath*}
    \UD{X}{Y} \hiderel{=}\int_{\mathbb{R}} f_X(x)\int_{\mathbb{R}}\vert f_{Y\vert X=x}(y)-f_Y(y)\vert dy dx.\label{eq: continuous \uo}
\end{dmath*}

Another formulation (more measure theoretical) would be:
\begin{dmath*}
    \UD{X}{Y}\hiderel{=}2\cdot\sup_{A\in \mathcal{B}(\mathbb{R}^2)} \left \{ \int_{A}(f_{X,Y}(x,y)-f_X(x)f_Y(y))dydx \right \} \label{eq: continuous mt}.
\end{dmath*}

\subsubsection{\bf Mix of discrete and continuous}
When $X$ is discrete and $Y$ is continuous, Equation~\ref{eq: general definition \uo} reduces to:
\begin{dmath*}
    \UD{X}{Y} \hiderel{=} \sum_{x} p_X(x)\int_{y}\vert f_{Y\vert X\hiderel{=} x}(y)-f_Y(y)\vert dy.  \label{eq: mix 1 \uo}
\end{dmath*}

Vice versa, if $X$ is continuous and $Y$ is discrete, Equation~\ref{eq: general definition \uo} becomes:
\begin{dmath*}
    \UD{X}{Y} \hiderel{=} \int_{x} f_X(x)\sum_{y}\vert p_{Y\vert X=x}(y)-p_Y(y)\vert   dx \label{eq: mix 2 \uo}.
\end{dmath*}

\subsection{Properties \uo (UD)} \label{appendix: properties \uo}
In this appendix, we prove properties of UD that are used in Section~\ref{sec: proof properties new dependency measure} to show that the {\it BP} dependency measure satisfies all properties in Table~\ref{tab: revised desirable properties}.

\subsubsection{\bf Symmetry UD} \label{sec: appendix symmetry UD}
UD is symmetric i.e. $\UD{X}{Y} = \UD{Y}{X}$ for every $X,Y$ as
\begin{dmath*}
\UD{X}{Y}=2 \sup_{A\in\mathcal{E}_X\bigotimes\mathcal{E}_Y} \left \{ \mu_{(X,Y)}(A)-(\mu_X\times\mu_Y)(A) \right \} = 2 \sup_{A\in\mathcal{E}_Y\bigotimes\mathcal{E}_X} \left \{\mu_{(Y,X)}(A)-(\mu_Y\times\mu_X)(A)\right \}=\UD{Y}{X}.
\label{eq appendix: symmetricity \uo}
\end{dmath*}

\subsubsection{\texorpdfstring{\bf Independence and UD $\bm{= 0}$}{}} \label{appendix: zero iff independent \uo}
Note that
\begin{dmath*}
\UD{X}{Y}=\sup_{A\in\mathcal{E}_X\bigotimes\mathcal{E}_Y} \left \{\mu_{(X,Y)}(A)-(\mu_X \times\mu_Y)(A) \right \}\\
+\sup_{B\in\mathcal{E}_X\bigotimes\mathcal{E}_Y} \left \{(\mu_X \times\mu_Y)(B)-\mu_{(X,Y)}(B) \right \}\\
\geq \left (\mu_{(X,Y)}(E_X\times E_Y)-(\mu_X \times\mu_Y)(E_x\times E_Y) \right ) 
+\left((\mu_X \times\mu_Y)(E_X\times E_Y)-\mu_{(X,Y)}(E_X\times E_Y) \right )\\
=0,
\end{dmath*}
with equality if and only if $\mu_{(X,Y)}=\mu_X\times\mu_Y$ on $\mathcal{E}_X\bigotimes\mathcal{E}_Y$, so if and only if $X$ and $Y$ are independent.\\

\subsubsection{\texorpdfstring{\bf Upper bound given $\bm{Y}$}{}} \label{appendix: proof maximality \uo}
To scale the dependency function, we need to determine $\sup_{X'}\left \{\UD{X'}{Y}\right \}$ for a given $Y$. Let $d_Y=\{y\in E_Y\vert \mu_Y(\{y\})>0\}$ be the set of all singletons with positive probability and let $c_Y=E_Y\setminus d_Y$. Furthermore, let $D_Y=E_X\times d_Y$ and $C_Y=E_X\times c_Y$. Then,
\begin{dmath*}
\UD{X}{Y}=2\sup_{A\in\mathcal{E}_X\bigotimes\mathcal{E}_Y} \left \{\mu_{(X,Y)}(A)-(\mu_X \times\mu_Y)(A) \right \}\\
=2\sup_{A\in\mathcal{E}_X\bigotimes\mathcal{E}_Y} \left \{\mu_{(X,Y)}(A\cap C_Y)-(\mu_X \times\mu_Y)(A\cap C_Y)\right \}\\
+2\sup_{A\in\mathcal{E}_X\bigotimes\mathcal{E}_Y} \left \{ \mu_{(X,Y)}(A\cap D_Y)-(\mu_X \times\mu_Y)(A\cap D_Y)\right \}.
\end{dmath*}

The first term is upper-bounded by
\begin{dmath*}
2\sup_{A\in\mathcal{E}_X\bigotimes\mathcal{E}_Y} \left \{ \mu_{(X,Y)}(A\cap C_Y)-(\mu_X \times\mu_Y)(A\cap C_Y) \right \}
\leq 2\mu_{(X,Y)}(C_Y)-0=2\mu_Y(c_Y)=2(1-\sum_{y\in d_Y}\mu_Y(\{y\})),
\end{dmath*}
with equality if and only if there exists a set $A$ such that ${\mu_{(X,Y)}(A\cap C_Y)=\mu_Y(c_Y)}$ and ${\mu_X \times \mu_Y (A\cap C_Y)=0}$.

The second term is upper-bounded by
\begin{dmath*}
2\sup_{A\in\mathcal{E}_X\bigotimes\mathcal{E}_Y} \left \{ \mu_{(X,Y)}(A\cap D_Y)-(\mu_X \times\mu_Y)(A\cap D_Y)\right \} =
2\sum_{y\in d_Y}\sup_{B\in\mathcal{E}_X} \left \{ \mu_{(X,Y)}(B\times \{y\})-(\mu_X \times\mu_Y)(B\times \{y\}) \right \}
\leq 2\sum_{y\in d_Y}\sup_{B\in\mathcal{E}_X}\left \{ \min\left \{ \mu_X(B),\mu_Y(\{y\})\right \}-\mu_X(B) \cdot \mu_Y(\{y\}) \right \}
= 2\sum_{y\in d_Y}\sup_{B\in\mathcal{E}_X}\left \{ \min\left \{\mu_X(B)\cdot (1-\mu_Y(\{y\})),(1-\mu_X(B)) \cdot \mu_Y(\{y\})\right \}\right\}
\leq 2\sum_{y\in d_Y} \left ( \mu_Y(\{y\}) \hiderel{-} \mu_Y(\{y\})^2 \right ),
\end{dmath*}
with equality if and only if for all $y\in d_Y$ we have that $\sup_{B\in\mathcal{E}_X}\left \{\mu_X(B)=\mu_Y(\{y\})\right \}=\sup_{B\in\mathcal{E}_X}\left \{\mu_{(X,Y)}(B\times\{y\})\right \} $.

Combining these two upper-bounds gives
\begin{dmath*}
\UD{X}{Y}\leq 2(1-\sum_{y\in d_Y}\mu_Y(\{y\})) + 2\sum_{y\in d_Y} \left (\mu_Y(\{y\}) \hiderel{-} \mu_Y(\{y\})^2 \right )  =  2\left (1-\sum_{y\in d_Y} \mu_Y(\{y\})^2 \right ).
\end{dmath*}

\subsubsection{\texorpdfstring{\bf Functional dependence attains maximum UD}{}} \label{appendix: proof maximum UD achieved}
If ${Y=f(X)}$ for a measurable function ${f:E_X\to\mathbb{R}^m}$, let ${d_Y=\{y\in\mathbb{R}^m \vert \mathbb{P}(Y=y)>0\}}$ be the set of elements $y$ with positive probability (which is a countable set) and let ${c_Y=\mathbb{R}^m\setminus d_Y}$. Then for any ${\epsilon>0}$ there exists a partition ${T_{1,\epsilon},T_{2,\epsilon},\dots,T_{k_\epsilon,\epsilon}}$ of $c_Y$ such that ${\mathbb{P}(Y\in T_{i,\epsilon})<\epsilon}$, thus ${\mu_X(f^{-1}(T_{i,\epsilon}))=\mathbb{P}(X\in f^{-1}(T_{i,\epsilon}))<\epsilon}$. Now we define the set $${B_\epsilon=\left (\bigcup_{i=1}^{k_\epsilon} \left (f^{-1}\left (T_{i,\epsilon}\right )\times T_{i,\epsilon}\right ) \right ) \cup \left (\bigcup_{y\in d_y}\left (f^{-1}\left (\{y\}\right )\times\{y\}\right )\right )},$$ then
\begin{dmath*}
\UD{X}{Y}=2\sup_{A\in\mathcal{E}_X\bigotimes\mathcal{E}_Y} \left \{ \mu_{(X,Y)}(A)-(\mu_X \times\mu_Y)(A) \right \}
\geq \sup_{\epsilon} \left \{ 2(\mu_{(X,Y)}(B_{\epsilon})-(\mu_X \times\mu_Y)(B_{\epsilon}))\right \}
\geq \sup_{\epsilon} \left \{ 2(1-\sum_{i=1}^{k_{\epsilon}} (\mu_X(f^{-1}(T_{i,\epsilon}))\mu_Y(T_{i,\epsilon}))-\sum_{y\in d_Y}\mu_X(f^{-1}(\{y\}))\mu_Y(\{y\})) \right \}
\geq \sup_{\epsilon} \left \{ 2(1-\epsilon\mu_Y(c_Y)-\sum_{y\in d_Y}\mu_Y(\{y\})^2) \right \}
=2(1-\sum_{y\in d_Y}\mu_Y(\{y\})^2).
\end{dmath*}

In Appendix~\ref{appendix: proof maximality \uo}, we have determined the upper bound, which is also equal to $2(1-\sum_{y\in d_Y}\mu_Y(\{y\})^2)$. Thus, UD is maximized. As corollary to this result, we find that ${\UD{Y}{Y}=0}$ iff there exists $y$ such that ${\mathbb{P}(Y=y)=1}$ so iff $Y$ is almost surely constant.

Note that the only property of $Y$ being constrained to $\mathbb{R}^m$ we really need is the fact that all atoms can be split into singletons and a null-set. So specifically, once we remove singletons with positive probability we are left with a non-atomic measurable space. This therefore allows us to define partitions. The proof can therefore be easily transferred to any space where this condition holds. 

Only for highly specific situations, we were unable to prove that the upper bound is tight and achieved by $\UD{Y}{Y}$. It specifically concerns the cases where there exist sets $A$ with $\mu_Y(A)>0$ and for all subsets $A'\subset A$ it holds that $\mu_Y(A')\in\{\mu_Y(A),0\}$ and additionally for all elements $a\in A$ it holds that $\mu_Y(\{a\})=0$. We call these 'non-trivial atoms'. In practice non-trivial atoms are highly irregular. It is mostly interesting from a theoretical point of view (for the sake of completeness).

We conjecture that for this case the upper bound would be equal to $\UD{X}{Y}\leq 2\left (1-\sum_{A\in B}\mu_Y(A)^2\right )$ where $B$ is a set of atoms of $\mu_Y$ with one representative per equivalence class (where $A_1\sim A_2$ if they differ by a null-set) and that this bound is attained for $X$ for which $Y=f(X)$ (so in particular $X=Y$). However, as mentioned above this problem remains open.

\mbox{}\\
\subsubsection{\texorpdfstring{\bf Measurable functions never increase UD}{}}
\label{appendix: never increase \uo}
Let ${f:(E_X,\mathcal{E}_X)\to (E_{X'},\mathcal{E}_{X'})}$ be a measurable function. Then ${h:E_X\times E_Y\to E_{X'}\times E_{Y}}$ with ${h(x,y)=(f(x),y)}$ is measurable. Now it follows that
\begin{dmath*}
\UD{f(X)}{Y}=2\sup_{A\in\mathcal{E}_{X'}\bigotimes\mathcal{E}_Y} \left \{\mu_{(f(X),Y)}(A)-(\mu_{f(X)} \times\mu_{Y})(A) \right \}
=2\sup_{A\in\mathcal{E}_{X'}\bigotimes\mathcal{E}_Y}\left \{\mu_{(X,Y)}(h^{-1}(A))-(\mu_{X} \times\mu_{Y})(h^{-1}(A)) \right \},
\end{dmath*}
with ${h^{-1}(A)\in \mathcal{E}_X\bigotimes\mathcal{E}_Y}$. Thus,
\begin{dmath*}
\UD{f(X)}{Y}\leq 2\sup_{A\in\mathcal{E}_X\bigotimes\mathcal{E}_Y}(\mu_{(X,Y)}(A)-(\mu_X \times\mu_Y)(A))=\UD{X}{Y}.
\end{dmath*}
In Appendix~\ref{sec: appendix symmetry UD}, it is proven that UD is symmetric. Therefore, it also holds for ${g:E_Y\to E_{Y'}}$, that
\begin{align*}
    \UD{X}{g(Y)}\leq \UD{X}{Y}.
\end{align*}

\mbox{}\\
\subsubsection{\texorpdfstring{\bf UD invariant under isomorphisms}{}}\label{appendix: invariant isomorphisms \uo}
Using Appendix~\ref{appendix: never increase \uo}, it must hold for all isomorphisms $f,g$ that
\begin{dmath*}
\UD{X}{Y}=\UD{f^{-1}(f(X))}{g^{-1}(g(Y))}\leq \UD{f(X)}{g(Y)}\leq \UD{X}{Y}.
\end{dmath*}
Therefore, all inequalities are actually equalities. In other words,
\begin{align*}
    \UD{f(X)}{g(Y)}=\UD{X}{Y}.
\end{align*}

\bibliographystyle{APT}

\end{document}